\documentclass[pmlr,twocolumn,10pt]{jmlr} % W&CP article

\usepackage{mathtools}

\usepackage{graphicx}

\usepackage[justification=centering]{caption}
\usepackage{color}
\usepackage{rotating}
\definecolor{mygreen}{gray}{0.8}
\usepackage{makecell}
\usepackage{xcolor}
\usepackage{multirow}
\usepackage{float}
\usepackage{longtable}
\usepackage{booktabs,colortbl}  
\definecolor{rowgray}{gray}{0.9}
\usepackage{amssymb}% http://ctan.org/pkg/amssymb
\usepackage{pifont}% http://ctan.org/pkg/pifont
\newcommand{\cmark}{\ding{51}}%
\newcommand{\xmark}{\ding{55}}%

\usepackage{booktabs}
\usepackage{siunitx}

\theorembodyfont{\upshape}
\theoremheaderfont{\scshape}
\theorempostheader{:}
\theoremsep{\newline}

\jmlrvolume{LEAVE UNSET}
\jmlryear{2022}
\jmlrsubmitted{LEAVE UNSET}
\jmlrpublished{LEAVE UNSET}
\jmlrworkshop{Conference on Health, Inference, and Learning (CHIL) 2022} % W&CP title

\title[MedMCQA : Multi-Subject Multi-Choice Dataset]{MedMCQA : A Large-scale Multi-Subject Multi-Choice Dataset for Medical domain Question Answering}

\author{%
 \Name{Ankit Pal} \Email{ankit.pal@saama.com}\\
 \Name{Logesh Kumar Umapathi} \Email{logesh.umapathi@saama.com}\\
 \Name{Malaikannan Sankarasubbu} \Email{malaikannan.sankarasubbu@saama.com}\\
 \addr Saama AI Research Chennai, India
}

\begin{document}

\maketitle

\begin{abstract}
This paper introduces MedMCQA, a new large-scale, Multiple-Choice Question Answering (MCQA) dataset designed to address real-world medical entrance exam questions. More than 194k high-quality AIIMS \& NEET PG entrance exam MCQs covering 2.4k healthcare topics and 21 medical subjects are collected with an average token length of 12.77 and high topical diversity. Each sample contains a question, correct answer(s), and other options which requires a deeper language understanding as it tests the 10+ reasoning abilities of a model across a wide range of medical subjects \& topics. A detailed explanation of the solution, along with the above information, is provided in this study.
\end{abstract}

\paragraph*{Data and Code Availability}
The dataset to reproduce these experiments and the leaderboard to track the progress of MedMCQA is available at \url{medmcqa.github.io}

\section{Introduction}
Question Answering (QA) is an important and challenging research area in Natural Language Processing (NLP). QA systems enable efficient
\begin{figure}
    \centering
  \includegraphics[width=4.8 cm]{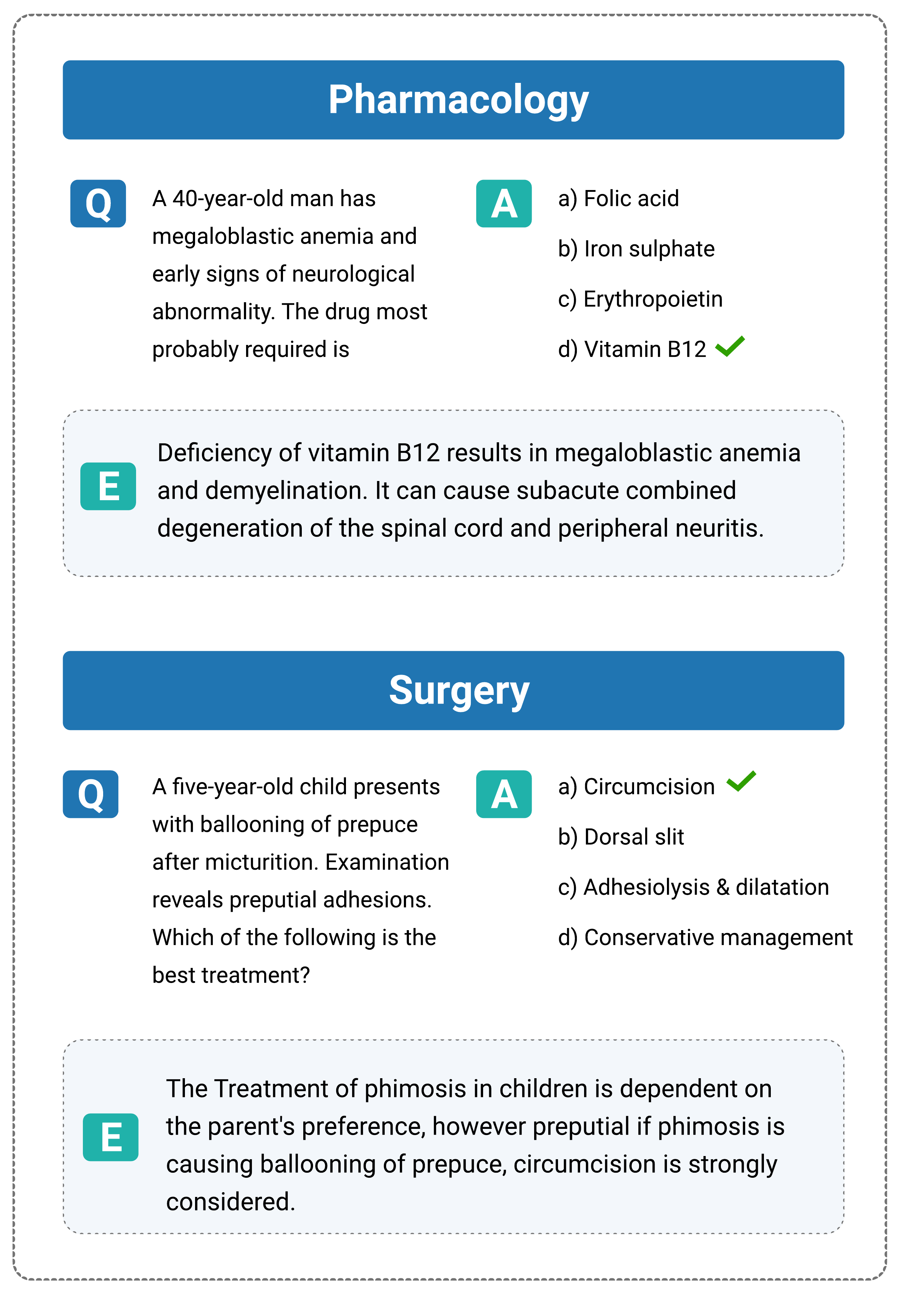} \caption{\footnotesize Samples from the MedMCQA dataset, along with the answer’s explanation.  ({\cmark} : the correct answer)}
  \label{fig:questions}
\end{figure}
access to the vast amount of information available that exists in text format.

In recent times, a significant amount of work has been done on constructing a question-answer dataset \citep{Rajpurkar2016,rajpurkar2018know,Reddy2019CoQAAC,Kwiatkowski2019,Yang2015} reading comprehension datasets \citep{Yang2018,Lai2017,Zellers2018,Yagcioglu2018,Dua2019,Bajaj2018,Huang2019}, extractive question answering \citep{Hermann2015,Trischler2017}, healthcare domain QA \citep{Jin2019,Clicr,CovidQA} and the organization of workshops \& competitions such as the Question Answering in the medical domain \& BioASQ Challenge \citep{MEDIQA, Nentidis_2020}

However, despite these successful efforts, automatic questions answering for real medical examination is still a challenge that is less explored. This type of real-world examination dataset on complex medical subjects like pharmacology, medicine, surgery, etc., is scarce. Apart from their scarcity, the requirement of a comprehensive understanding of the domain, matching human experts, makes them appealing for research pursuits.
Before this attempt, very few works have been done to construct biomedical MCQA datasets \citep{Vilares2019}, and they are (1) mostly small, containing up to few thousand questions, and (2) cover a limited number of Medical topics and Subjects.

Thus, a large-scale, diverse medical QA dataset is needed to accelerate research and facilitate more consistent and effective open-domain QA models in Medical-QA. This paper addresses the aforementioned limitations by introducing MedMCQA, a new large-scale, Multiple-Choice Question Answering (MCQA) dataset designed to address real-world medical entrance exam questions. The dataset consists of 194k high-quality medical domain MCQs covering 2.4k healthcare topics and 21 medical subjects to provide a reliable and diverse benchmark. Apart from the question, the correct answer(s), and other options., it also consists of various ancillary data, the primary being a detailed explanation of the solution.

Questions are taken from AIIMS \& NEET PG entrance exam MCQs, where graduate medical students are evaluated on their professional knowledge. Questions in these exams are challenging and generally require deeper domain and language understanding as it tests the 10+ reasoning abilities across a wide range of medical subjects \& topics. Hence a model must be trained to find relevant information from the open domain knowledge base, reason over them, and choose the correct answer.

Fig.\ref{fig:questions} shows two example questions, their corresponding explanation, and answers from the study dataset.

An in-depth analysis \& a thorough evaluation of the dataset are conducted. The baseline experiments on this dataset with the current state-of-the-art methods can only answer 47\% of the question correctly, which is far behind the performance of human candidates (merit candidates of these exams score an average of 90\% marks). Error analysis and results indicate possibilities for improvement in the current methods' reasoning and medical domain question answering. It is believed that this dataset would be an appropriate testbed for future research in this direction.

In brief, the contributions of this study are as follows.

\begin{itemize}
    \item \textbf{Diversity and difficulty} This dataset offers several advantages over existing datasets: (i) Covers ~2.4k healthcare topics and 21 medical subjects with an average token length of 12.77, the diversity of questions in MedMCQA demonstrate challenges unique to the dataset. (ii) It is larger than pre-existing Medical QA datasets, (iii) As these questions are from real-world and mock examinations, all the questions and candidate options are created by human experts. These questions are a comprehensive evaluation of a medical practitioner's professional skills, (iv) The questions are difficult \& challenging. They test the 10+ reasoning abilities of a model across a wide range of medical subjects \& topics.
    
    \item \textbf{Quality} Detailed statistics, analysis of the data, and fine-grained evaluation per medical subject are provided, yielding a more precise comparison between models. Each sample contains a question, correct answer(s), other options, and a detailed explanation of the solution.
    \item \textbf{Evaluation of quality} Extensive experiments are conducted using high-performance pre-trained medical domain models. Error analysis is also provided to illustrate the major challenges of this task. The baseline experiments on this dataset with the most current state-of-the-art methods answer only 47\% of the question correctly, which is far behind the human performance of 90\%, indicating possibilities for improvement in models' reasoning ability \& constitutes a challenging benchmark for future research.

    \item \textbf{Reproducible exam-based split} The dataset is split based on the exams instead of a question-based split (explained in section \ref{apd:gll}). This ensures that the evaluation is closer to the real-world examinations, model generalizability, and reusability. Individual Examinations tend to have similar questions or pattern of questions repeated periodically. Exam based split avoid this leakage of similar questions into test set, hence helping in generalizability of the dataset. The dataset code to reproduce the experiments \& the leaderboard to track the progress of MedMCQA are available at \url{medmcqa.github.io}

\end{itemize}

\begin{table*}[!ht]
\small
\centering
\begin{tabular}{lccccccc}
\toprule
 {\bf Dataset} & {\bf \# Question} & {\bf \# Subject} & {\bf  Publicly Available} & {\bf Explanation}  & {\bf Split Type} & {\bf Open Domain} \\
\midrule
MedQA    &  270,000 &   -   &  {\xmark}  &  {\xmark}  &  random  &  {\cmark}  \\
HEAD-QA  &  13,530   &  6   &  {\cmark}  & {\xmark}  &  yearwise   &  {\cmark}  \\

\textbf{MedMCQA}  &  193,155 & 21  & {\cmark}  & {\cmark} & exam-based  &  {\cmark}  \\
\bottomrule
\end{tabular}
\caption{Comparison of MedMCQA with several existing MCQA datasets(MedQA\citep{zhang2018medical}, HEAD-QA\citep{Vilares2019}) in the medical domain. {\cmark} represents the dataset that has the feature and {\xmark} represents it does not}
\label{tab:comparison}
\vspace{-2ex}
\end{table*}

\section{The MedMCQA Dataset}

In this section, properties of the MedMCQA dataset are presented. Data collection, preparation, preprocessing, and train/test/development splits are discussed.

\subsection{Task Definition}
The MedMCQA task can be formulated as $\mathbf{X = \{Q, O\}}$ where $\mathbf{Q}$ represents the questions in the text, $\mathbf{O}$ represents the candidate options, multiple candidate answers are given for each question $\mathbf{O = \{O_{1}, O_{2}, ..., O_{n} \}}$. The goal is to select the single or multiple answers from the option set.The ground truth label of a data point is $y \in \mathbb{R}^{n}$ where ${y}^{i}$ $=$ $\mathbf{\{0,1\}}$ and ${n}$ is the number of options, the objective is to learn a prediction function ${f : X \rightarrow {y}}$

\subsection{Dataset collection}

All India Institute of Medical Sciences (AIIMS PG) \& National Eligibility cum Entrance Test (NEET PG) are the two medical entrance exams conducted by All India Institute for Medical Sciences (AIIMS) \& National Board of Examinations (NBE), respectively, for providing admission to the postgraduate medical courses. The applicants must have obtained an Bachelor of Medicine and Bachelor of Surgery (MBBS) from a recognized institute to appear for the exams. The exams are used to evaluate the candidates in a structured format, namely, Diagnostic Reasoning and Treatment, Pharmacology, Psychology, Biology, Physical Examination, General Management Strategies, Medical Knowledge, and many other aspects of health and general attitude demeanor of the patient and the examiners. These exams are a comprehensive evaluation of the professional skills of a medical practitioner.

In this paper, the raw data is collected from open websites and books that put together several mock tests and online test series created by medical professionals. In addition to the collected data, AIIMS \& NEET PG examination questions (1991- present) from the official websites are also used to create the MedMCQA. 

The dataset contains MCQs with fine-grained human-labeled classes on various graduation level medical subjects. Each sample contains ID, question, correct answer, and options. Besides, an explanation of the solution is also provided.

\subsection{Preprocessing \& Quality Checks}
To ensure that all the questions are answerable using textual input only,
the following steps were taken to clean the raw data, considering questions from several data sources, 

\begin{itemize}
    \item Questions with an inconsistent format were excluded, e.g., a question where the number of options was not four(excluding punctuation marks).
    \item Questions with no best answer and missing or null candidates were also omitted.
    \item Questions whose validity relied on external information were filtered, i.e., the articles and questions containing images or tables.
    \item Questions containing the keywords ``equation", ``India", ``graph", ``map" etc., were removed using a manually curated list of words.
    \item Further, heuristic rules were also used. For example, in some cases, the question contained HTML tags, special symbols, URLs, extra whitespaces, and missing options. Different tools were used, e.g., a spell checker, an HTML parser, to identify and correct these cases.
    \item A proofreading tool, `Grammarly' was used for all the questions, options, and explanations in the dataset to fix the grammar, punctuation, and spelling mistakes. Appropriate suggestions from the tool were applied to the content with human supervision to improve the dataset's quality. As a result, many errors could be corrected
    \item Lastly, all duplicated questions were removed.
\end{itemize}

% Firstly, all questions with an inconsistent format are excluded, e.g., a question would be excluded if the number of options is not four (excluding punctuation marks). Questions with no best answer and missing or null candidates were also omitted.

% Also, all questions are filtered whose validity relies on external information, i.e., the articles and questions containing images or tables are removed. all questions containing the keywords “equation,” “India,” “graph,” “map,” etc., are removed using a manually curated list of words.

% Further, heuristic rules are also used to clean the data. For example, in some cases, the question contains HTML tags, special symbols, URLs, and whitespaces between two missing words.  different tools were used, e.g., spell checker, Html parser, etc., to identify and correct these cases. Besides, we also run a proofreading tool, Grammarly, on all the questions, options, and explanations in the dataset to fix the grammar, punctuation, and spelling mistakes. Appropriate Suggestions from the tool are applied on the content with human supervision to improve the dataset contents' quality.

% Lastly, all duplicated questions are removed. The final dataset contains 192,555 questions.

% As a result, many errors could be corrected. Finally, additional experiments were carried out to ensure that the question has provided information that matches the predefined goals.
Additional data cleansing steps were carried out to ensure that the question has provided information that matches the data quality goals. The final dataset contains 193,155 questions.

\subsection{Split Criteria}
\label{apd:gll}
The goal of MedMCQA is to emulate the rigor of real word medical exams. To enable that, a predefined split of the dataset is provided. The split is by exams instead of the given questions. This also ensures the reusability and generalization ability of the models. 

The training set of MedMCQA consists of all the collected mock \& online test series, whereas the test set consists of all AIIMS PG exam MCQs (years 1991-present). The development set consists of NEET PG exam MCQs (years 2001-present) to approximate real exam evaluation.

In the dataset, leakages of similar questions from the training data to test and dev could artificially inflate the models' performance. This is avoided by building the development and test set to include sufficiently different training data questions. 

The Levenshtein distance between each pair of questions was computed in the entire dataset. If the similarity between the two documents was larger than 0.9, the question was excluded from the development and test set. The final dataset contains ~183K train examples, 6K in the development set, and 4K in the test set.

\begin{figure}
\centering
  \includegraphics[width=6 cm]{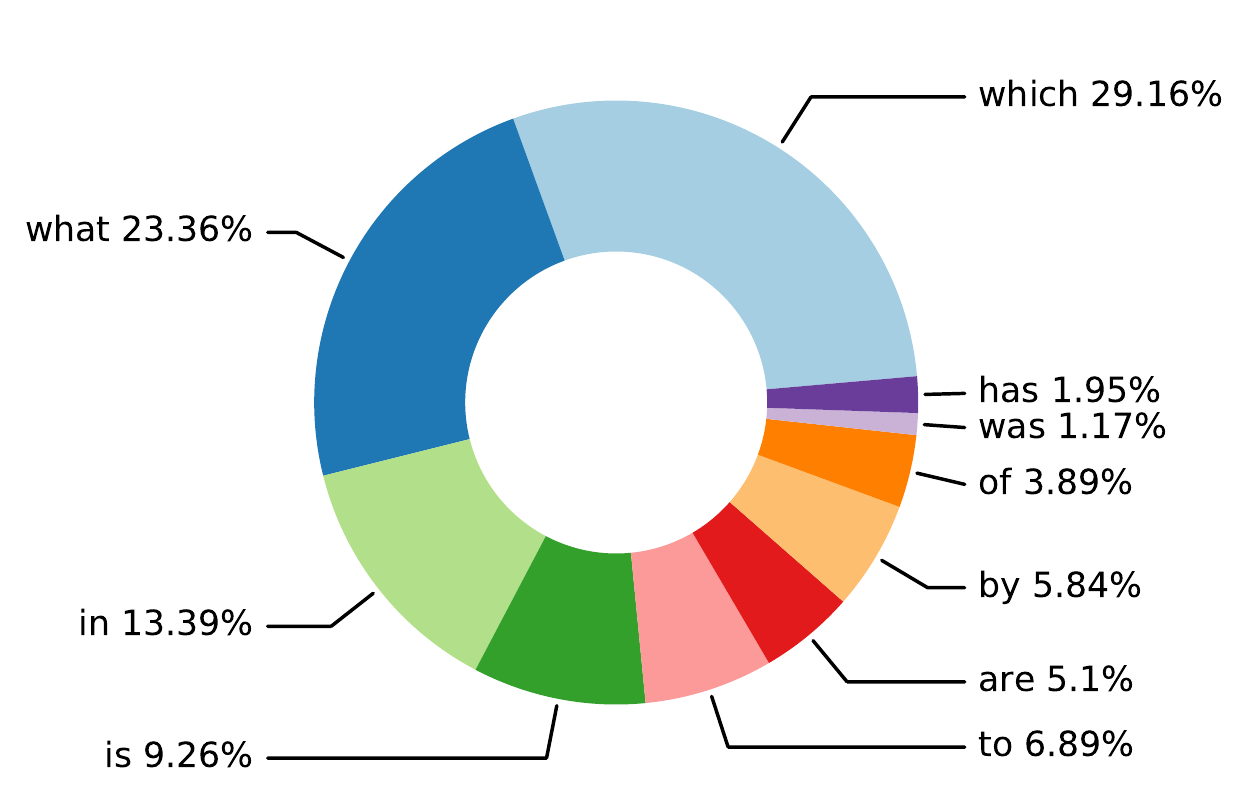}
  \caption{ \footnotesize Relative sizes of Question Types in MedMCQA}
  \label{fig:question_types}
\end{figure}

\begin{figure}[ht]
\centering
  \begin{minipage}[b]{0.50\linewidth}
   
    \includegraphics[width=0.90\linewidth]{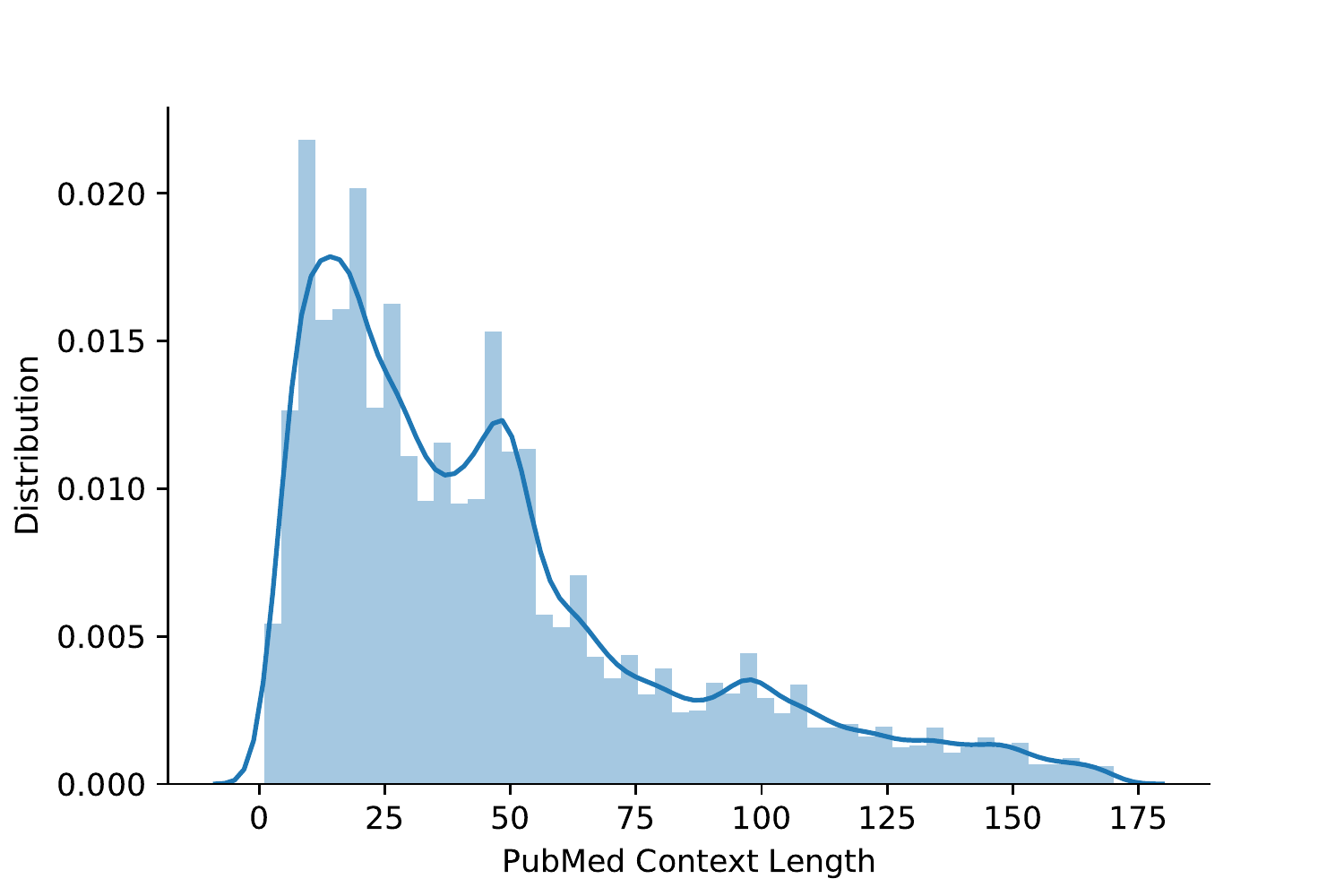}
    
    \label{fig:stasdet0}
    \vspace{2ex}
  \end{minipage}%%
  \begin{minipage}[b]{0.50\linewidth}

    \includegraphics[width=.90\linewidth]{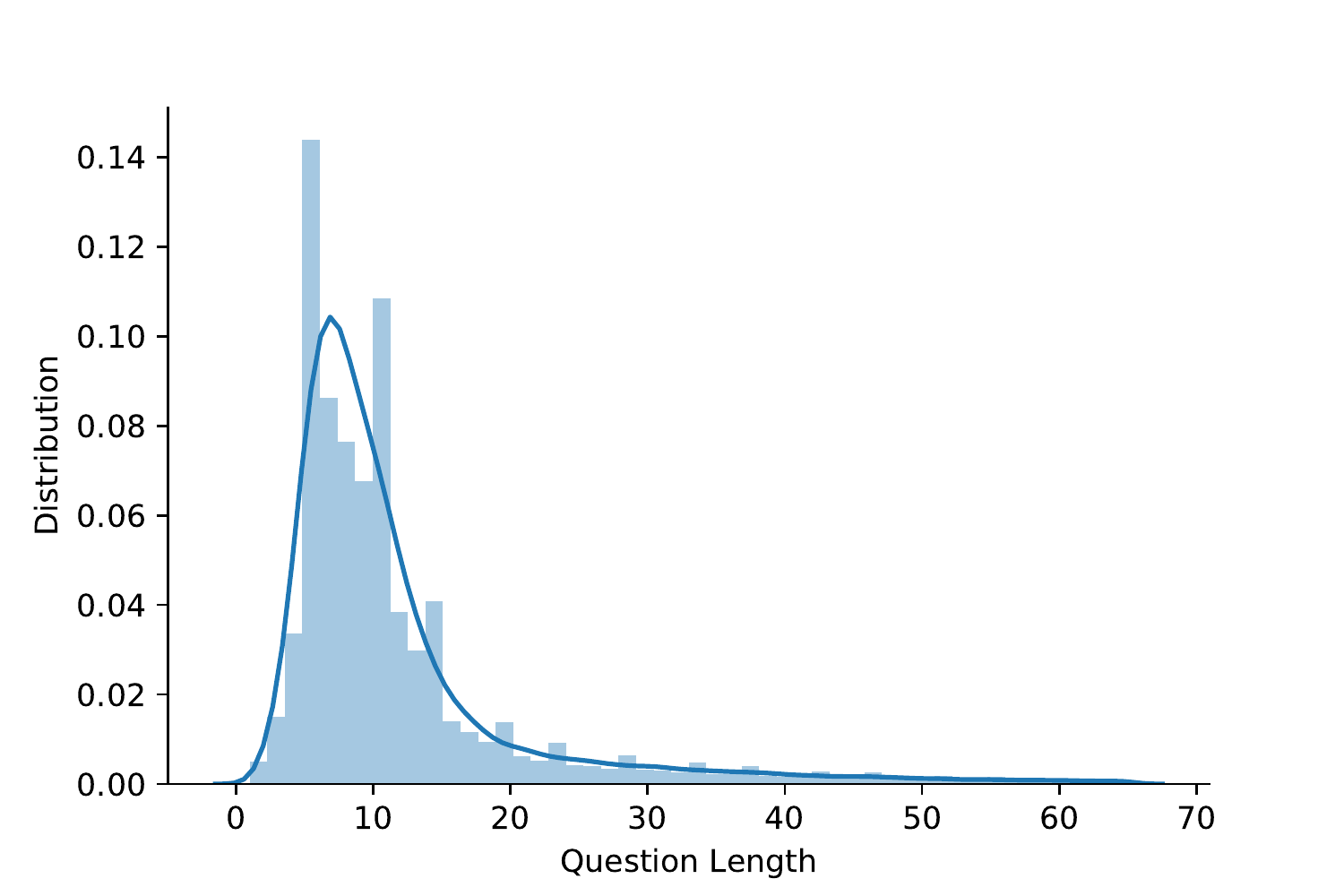} 
    \label{fig:stasdet1}
    \vspace{2ex}
  \end{minipage} 
  \begin{minipage}[b]{0.50\linewidth}

    \includegraphics[width=.90\linewidth]{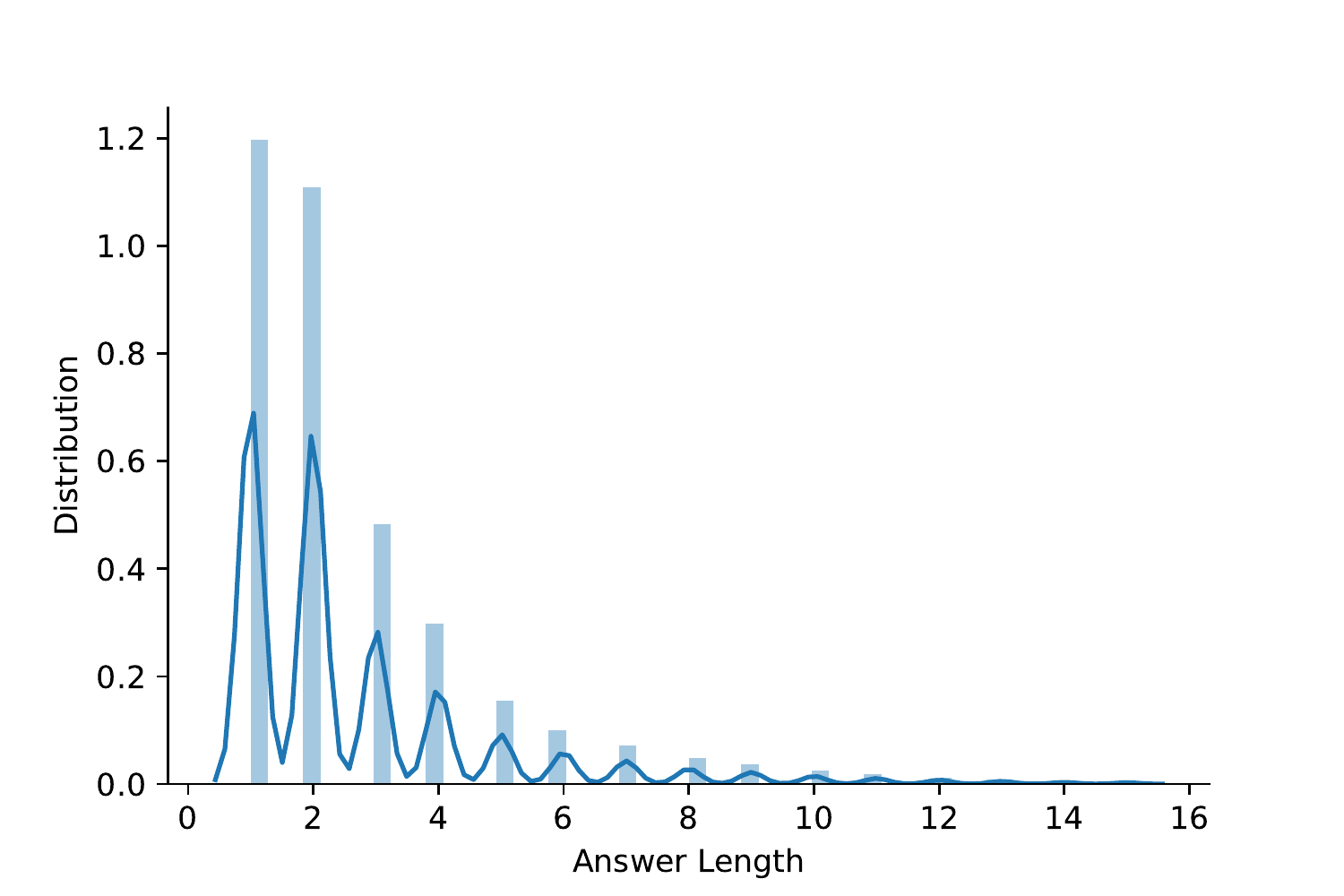} 
    \label{fig:stasdet2}
    \vspace{2ex}
  \end{minipage}%% 
  \begin{minipage}[b]{0.50\linewidth}

    \includegraphics[width=.90\linewidth]{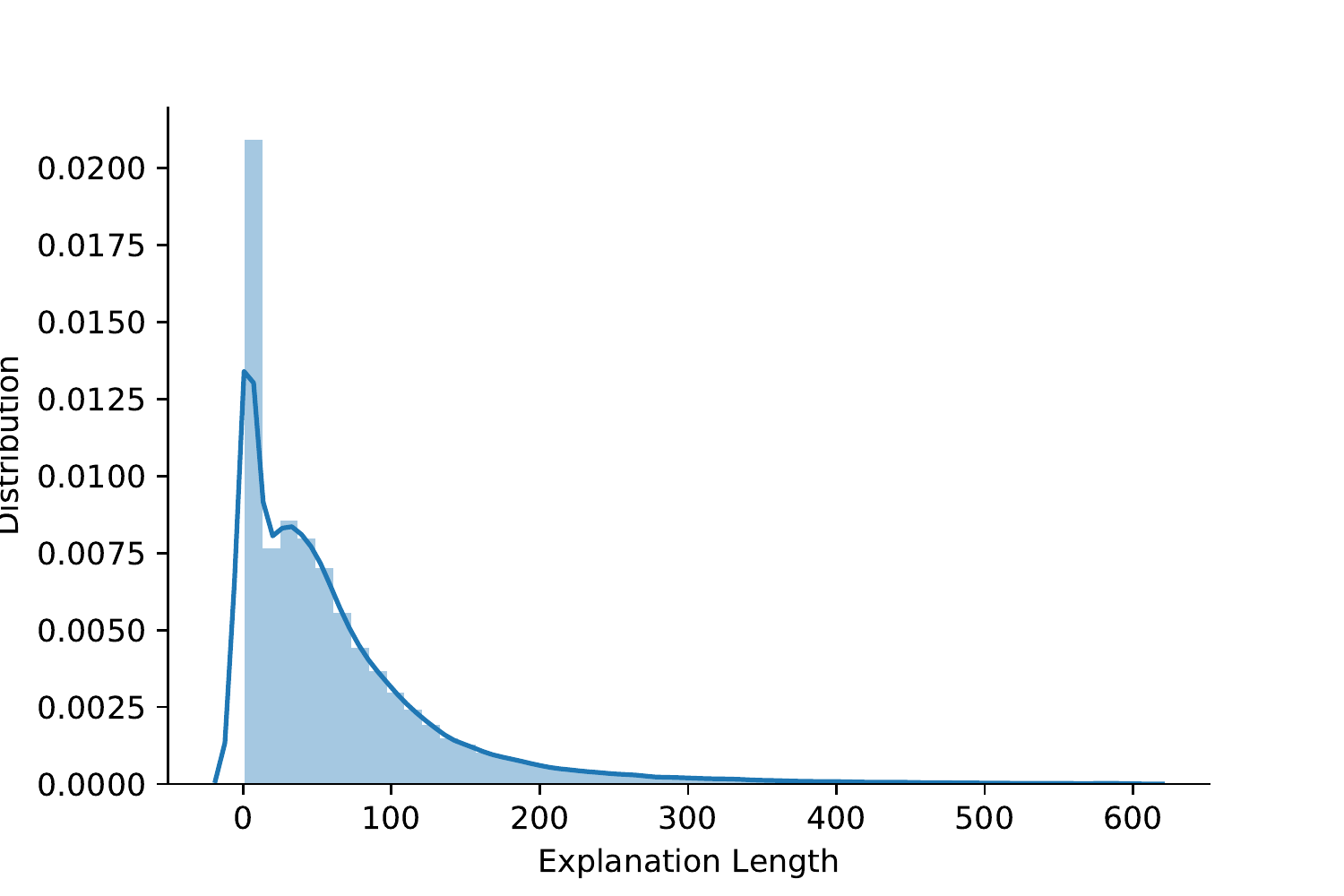}
    \label{fig:stasdet3}
    \vspace{2ex}
  \end{minipage}
  \caption{ \footnotesize (a) distribution of Pubmed context length (b) Distribution of question length (c) Distribution of answer length (d) Distribution of explanation length }
  \label{fig:stasdet}
\end{figure}

% \begin{figure}[ht]
%   \begin{subfigure}[b]{0.2} 
%     \includegraphics[width=\linewidth]{images/PubMed Context.pdf}
%     \caption{\footnotesize Context Length} 
%     \label{fig:stasdet0}   
%   \end{subfigure}
%   \hfill % maximize the horizontal separation
%   \begin{subfigure}[b]{0.2}
%     \includegraphics[width=\linewidth]{images/Question.pdf} 
%     \caption{\footnotesize Question Length}
%     \label{fig:stasdet1}   
%   \end{subfigure} 

%   \bigskip % provide some vertical separation
%   \begin{subfigure}[b]{0.2}
%     \includegraphics[width=\linewidth]{images/Answer.pdf} 
%     \caption{ \footnotesize Answer Length}
%     \label{fig:stasdet2}
%   \end{subfigure}
%   \hfill  % maximize the horizontal separation
%   \begin{subfigure}[b]{0.2}
%     \includegraphics[width=\linewidth]{images/Explanation.pdf}
%     \caption{\footnotesize Explanation Length}
%     \label{fig:stasdet3} 
%   \end{subfigure}

%   \caption{(a) Distribution of Pubmed context length (b) Distribution of question length (c) Distribution of answer length (d) Distribution of explanation length}
%   \label{fig:stasdet}
% \end{figure}

\section{Data statistics}
This dataset covers many medical subjects based on the AIIMS \& NEET PG entrance exams. The train, development, and test set consist of 182,822 , 4,183 \& 6,150 questions with an average token length of 12.35, 13.91 \& 9.68, respectively.
The general statistics of preprocessed data are summarized in table \ref{tab:data_split}

\begin{figure*}[!ht]
  \includegraphics[width=16cm]{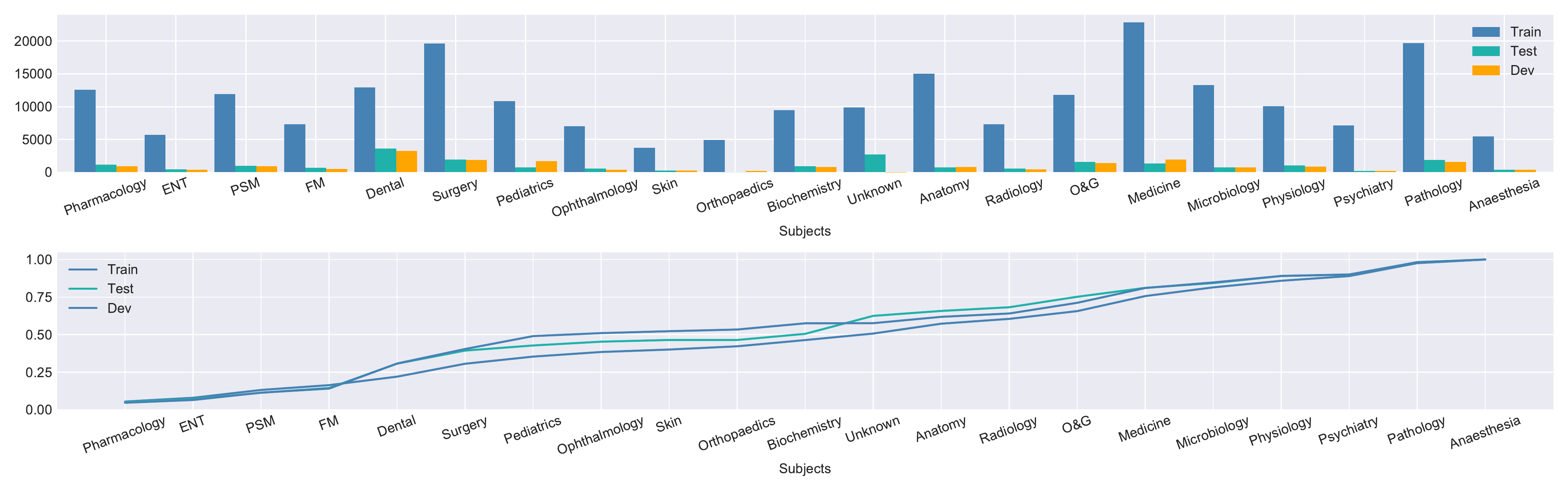}
  \caption{ \footnotesize Distribution of unique tokens \& Cumulative Frequency Graph in the union of Train, Test, and Development split in MedMCQA dataset. The vocabulary size in the AIIMS PG exams (Test Set) is larger than that of the NEET exams (Dev. Set). Thus indicating the correlation between vocabulary size and difficulty level of the exam. 
}
  \label{fig:token_dist}
\end{figure*}

An additional informative statistic is the count of unique tokens in the dataset plotted in Fig. \ref{fig:token_dist}. Vocabulary size is a good measure of linguistic and domain complexity associated with a text corpus and influences the models' performance. It is observed that the length of questions and the vocabulary size in the AIIMS PG exams (test set) are larger than that of the NEET PG exams (dev. set). Hence, it can be inferred that questions from AIIMS are more complex than NEET.

\begin{table}[!ht]
\footnotesize
\centering
\begin{tabular}{lcccc}
\toprule
 & {\bf Train} & {\bf Test} & {\bf Dev} & {\bf Total} \\
\midrule
Question \#  &  182,822 & 6,150  & 4,183  & 193,155 \\
Vocab & 94,231& 11,218 & 10,800 & 97,694 \\
Max Q tokens & 220 & 135 & 88 & 220  \\
Max A tokens & 38 & 21 & 25& 38 \\
Max E tokens & 3,155 & 651& 695& 3,155 \\
Avg Q tokens & 12.77 & 9.93 & 14.09& 12.71 \\
Avg A tokens & 2.69& 2.58& 3.19& 2.70 \\
Avg E tokens & 67.52 & 46.54 & 38.44& 66.22 \\
\bottomrule
\end{tabular}
\caption{MedMCQA dataset statistics, where Q, A, E represents the Question, Answer, and Explanation, respectively}
\label{tab:data_split}
\vspace{-2ex}
\end{table}

\section{Data Analysis}
An analysis of the dataset is presented in the subsequent sections. The difficulty and diversity of questions and the answers were analyzed to understand the MedMCQA dataset's properties. The complexity of MedMCQA is demonstrated by considering the question and reasoning types covered in the dataset. 

\subsection{Difficulty and Diversity of Questions}
In clinical medicine, a diverse number of questions are possible as it is spread over a range of topics. For example, given the description of a patient's condition, the question might be asked for the most probable diagnosis/the most appropriate treatment or examination required/mechanism of a certain condition, etc.

The majority of the dataset questions are non-factoid and open-ended in nature and seek detailed
information about the health condition. Questions in MedMCQA are fairly long, with a mean length of 12.77 words, indicating the compositional nature of questions and different levels of complexity and details covered.

To understand the types of questions in MedMCQA,  25\% of questions were sampled, and their properties were analyzed manually. It was observed that 68\% of the questions started with an interrogative word, which generally tends to be open-ended. The dataset also contained many dichotomous questions, which often require explanations. The diversity of questions in the MedMCQA makes it a challenging dataset containing many aspects of medical knowledge. Another distinguishing factor of this dataset is that it has questions that were created for and by human domain experts.

\subsection{Answer types}
In the dataset, each question contains four options with an average length of 2.69 tokens. Out of which, 25\% examples were sampled from the development set, and the answer types are presented in Fig. \ref{fig:answer_types}. As shown, MedMCQA covers a broad range of answer types, which matches the analysis on questions' contribution.
The answers were manually categorized, and it was observed that answers regarding drug/medicine's name accounted for 22.49\%. Medical procedure/Treatment type aiming to determine, measure, or diagnose a condition or parameter accounted for 18.74\% of answers. In comparison, 11.24\% of answers were related to the quantity of dose(in unit). It was observed that side effects, causes \& affected body parts accounted for 12.74\%, 10.49\% \& 9.75\% of the dataset. The rest of the answer groups contained fewer instances of the time period, adverse events \& other types.

\subsection{Subject \& Topic Analysis}
Fig. 8(\ref{apd:first}) in the Appendix presents the distribution of medical topics per subject for the datasets. Almost 95\% of the subjects contain above 50 topics, while 70\% of subjects exceed 100 topics exhibiting a plethora of medical content. Topics range from Medicine (Endocrinology, Infection, Haematology, Respiratory, etc.), Surgery (General Surgery, Endocrinology, breast, and Vascular surgery, etc.) to Radiology \& Biochemistry. This wide range of topics increases the dataset's difficulty.

\subsection{Reasoning Types}
\label{apd:re_type}
To provide a detailed \& better understanding of the dataset\'s reasoning types,  25\% of questions from MedMCQA were sampled randomly. The reasoning types required to answer were manually analyzed. The procedure was followed, and the annotation types presented in \citep{Clark2018} were re-used to categorize them into the following reasoning types:

\begin{figure}
  \includegraphics[width=7.5cm]{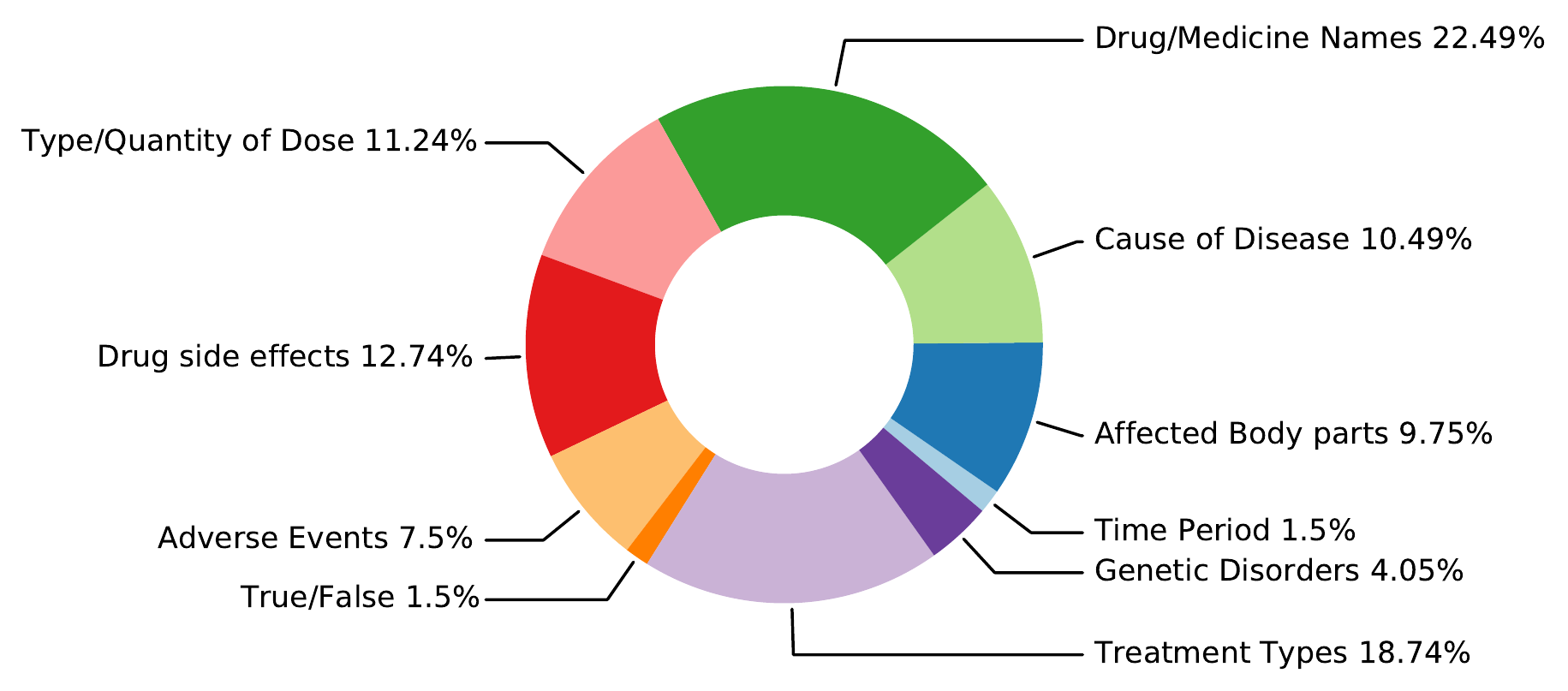}
  \caption{ \footnotesize Relative sizes of Answer Types in MedMCQA}
  \label{fig:answer_types}
\end{figure}

\begin{figure}
\centering
  \includegraphics[width=7 cm]{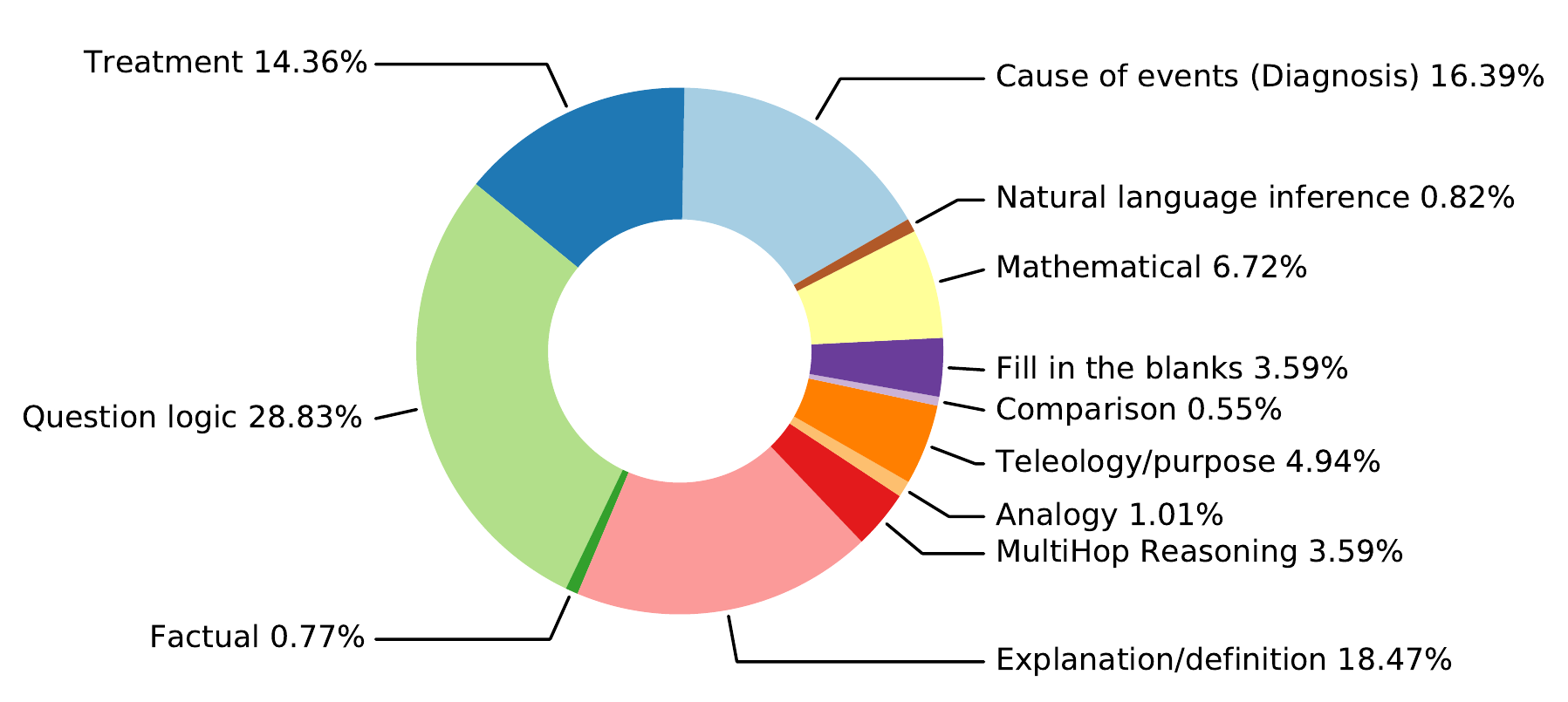}
  \caption{\footnotesize Relative sizes of Reasoning Types in MedMCQA}
  \label{fig:reasoning_types}
\end{figure}

\begin{itemize}
    \item \textbf{Question logic} In this, the reasoning is tested by excluding the distractor.
    \item \textbf{Factual} These are the questions that have facts as answers.
    \item \textbf{Explanation/definition} The questions that require selection of definition or explanation or a term/phenomenon.
    \item \textbf{MultiHop Reasoning} To answer these questions, the reasoning is required from multiple passages.
    \item \textbf{Analogy} In these types of questions, the responder must select the most similar/analogous answer.
    \item \textbf{Teleology/purpose} Requires understanding of the purpose of a phenomenon/a thing.
    \item \textbf{Comparison} Questions that require reasoning by comparing multiple options.
    \item \textbf{Fill in the blanks} The responder selects the most appropriate answer suitable to fill the blanks.
    \item \textbf{Natural language inference} Determining whether a hypothesis is true, false (contradiction), or neutral given an assumption.
    \item \textbf{Mathematical} Questions that require mathematical critical thinking and logical reasoning.
    \item \textbf{Treatment} Questions that require selection of a correct treatment method for a given ailment / condition.
    \item \textbf{Diagnosis} Questions that require selection of a correct cause of a given ailment / condition.

\end{itemize}
Fig. \ref{fig:reasoning_types} shows statistics \& examples of major reasoning types in the dataset.

% \begin{figure*}[!ht]
%   \includegraphics[width=16cm]{images/topicpersub1.pdf}
%   \caption{ \footnotesize Illustration}
%   \label{fig:maindiagram}
% \end{figure*}

\section{Baseline Models}
\label{apd:bm}
The primary motivation of the baseline experiments is to understand the adequacy of the current models in answering multiple-choice questions meant for human domain experts (post-graduate medical students) and to understand the level of domain specificity required in the models. Therefore, models and knowledge sources with varying levels of specificity are selected. We consider four existing models in our baseline experiments. 

They are based on different pre-trained language models using Transformers architecture \citep{NIPS2017_3f5ee243} , including BERT \citep{Devlin2019} , SciBERT \citep{Beltagy2019}, BioBERT \citep{c99d46c12d234e77957c3d847b64f5cf} and PubmedBERT\citep{Gu2020}. We fine-tuned these models on our training dataset in a multiclass classification fashion. We consider models of base size. BERT is evaluated for its out-domain pretraining, SciBERT and BioBERT for their mixed domain and in-domain continual training, and PubmedBERT for its in-domain pretraining. These models are explained in detail in the following section,

\subsection{SciBERT}
SciBERT \citep{Beltagy2019} is a pretrained language model based on BERT. The model has been pre-trained from scratch on 1.14M papers on the semantic scholar. Even though SciBERT has been pre-trained from scratch, it has a mix of computer science (18\%) and biomedical domain (82\%), making it a mix-domain pretrained model. The uncased version of the model that uses a vocabulary called scivocab is used, which is a domain-specific vocabulary of size 30K

\subsection{BioBERT}
BioBERT \citep{c99d46c12d234e77957c3d847b64f5cf} is the first biomedical domain-specific pretrained language model based on BERT. The model is initialized with standard BERT weights  (pretrained from Wikipedia and BookCorpus), and continual pretraining is performed with PubMed abstracts and full texts. The model uses the same vocabulary as the standard BERT model. The base variant of the 1.1 version of the model is used in the experiments.
\subsection{PubMedBERT}
PubMedBERT \citep{Gu2020} is a recent domain-specific pre-trained language model that is first to pretrain only on in-domain texts (PubMed abstracts

\begin{figure*}
 \centering
  \includegraphics[width=10cm]{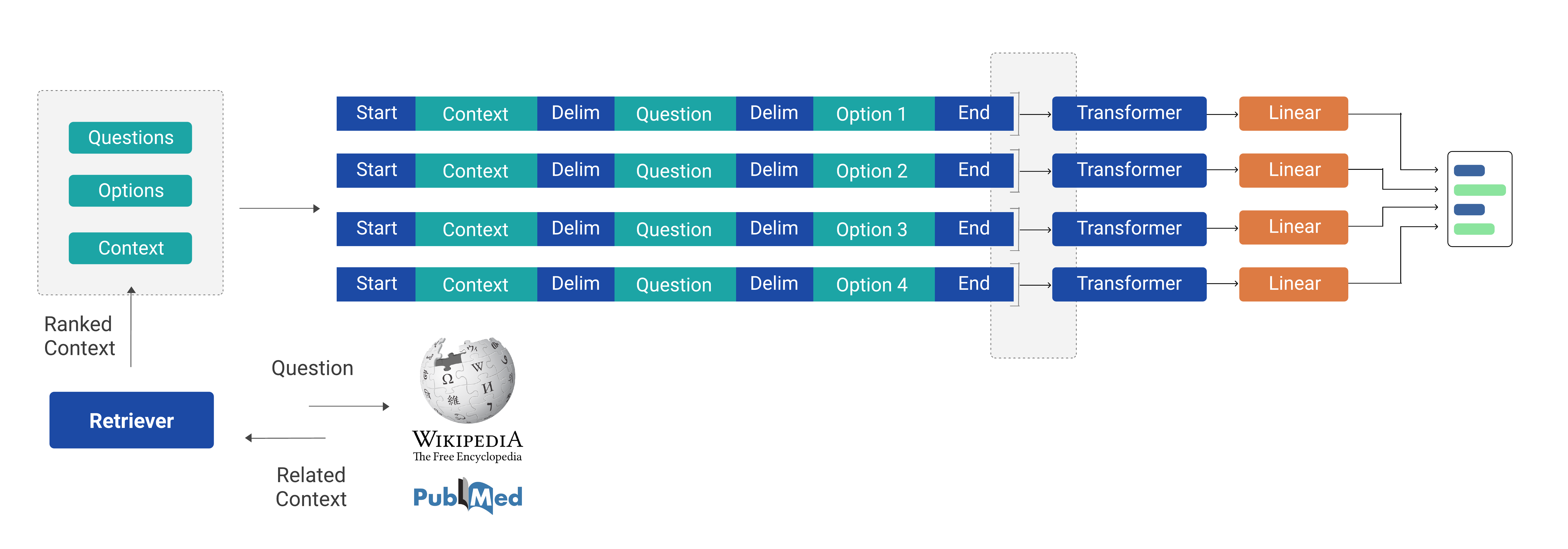}
  \caption{ \footnotesize The Retriever+Reader Pipeline for Open-Domain Question Answering system used in our experiments.Dense passage retrieval \citep{karpukhin2020dense} and PubMedBERT \citep{Gu2020} are used to evaluate Wikipedia and PubMed as knowledge bases respectively, while different transformer models (explained in section \ref{apd:bm}) as reader models.}
  \label{fig:open_domain}
\end{figure*}

\noindent and full texts). The base version of the model trained with both abstracts and full texts is used in the experiments. This model is used to evaluate the performance of a fully in-domain pre-trained model on the dataset.
\subsection{Retriever models}
With the recent success of neural retrievers, dense passage retrieval \citep{karpukhin2020dense}, and PubMedBERT\citep{Gu2020} were utilized to evaluate Wikipedia and PubMed as knowledge bases, respectively. Dense passage retriever follows a siamese/bi-encoder architecture; One encoder encodes the documents and another to encode the query, originally trained with Maximum inner product search objective. The pretrained DPR model and Wikipedia index from Transformer's library \citep{Wolf2020} were used in the experiments.
\section{Experiments}
\label{apd:exp}
To complement the motivation stated in section \ref{apd:bm}, The reader models were chosen with varying domain specificity levels. The contribution of external knowledge sources (Wikipedia and PubMed) was evaluated by providing these sources as contexts. Furthermore, an ablation study was also performed on context by training and evaluating all the models without context. This was done to understand the contribution of external context and the usefulness of the internal knowledge stored in these domain-specific models. The baseline experiments are broadly classified as follows,
% We can broadly classify the baseline experiments into the below categories,

\begin{itemize}
    
    \item \textbf{Out-Domain}:
    Pre-trained models trained on out-domain corpora like Wikipedia and Book corpus were used in this experiment type.
    
    \item \textbf{Mix domain (continual)}:
     Pre-trained models trained on out-domain initially and later adapted to in-domain or trained from scratch on both out-domain and in-domain corpora were used in this experiment.
     
    \item \textbf{In-Domain}:
    Pre-trained models trained from scratch on in-domain corpora like PubMed abstracts and full texts were used in this experiment type.
    
\end{itemize}

All these experiments were repeated with and without external knowledge context.

\subsection{Pubmed Data Preprocessing}
Before encoding the passages, the passages were truncated to 250 token lengths to fit the memory.

\subsection{Retriever}
\label{apd:rt}
For the experiments that involve context, a retriever+reader pipeline approach was opted  (as introduced in \citep{Chen2017}). The out-of-the-box retriever models were used  (explained in the section \ref{apd:rt}) from Huggingface's Transformers library \citep{Wolf2020} to encode the passages and questions. The passage with the highest cosine similarity was retrieved and used as a context for training the reader models.

\subsection{Reader finetuning}
The finetuning approach was followed as in \citep{Devlin2019} to finetune the reader models. The highest scoring contexts for each question are retrieved from the retriever. These contexts are combined by \colorbox{mygreen}{ \footnotesize  [SEP] } token with the concatenation of question and answer pair. This creates four input sequences per question.

\colorbox{mygreen}{ \footnotesize [CLS] }Context\colorbox{mygreen} { \footnotesize [SEP] }Question\colorbox{mygreen}{ \footnotesize  [SEP] } Option\colorbox{mygreen}{ \footnotesize [SEP] }

A linear layer with softmax is applied over the output of the \colorbox{mygreen}{ \footnotesize  [CLS]} token of the encoder. This is to select the most appropriate option for a question and context pair.

For the experiments that do not use context, question and answer pair concatenation is encoded, and a linear layer with softmax is applied over the output of the \colorbox{mygreen}{ \footnotesize  [CLS] } token of the encoder to select the most appropriate option for a question.

\colorbox{mygreen}{ \footnotesize [CLS] } Question \colorbox{mygreen}{ \footnotesize  [SEP] } Option \colorbox{mygreen}{ \footnotesize [SEP] }

The models were finetuned on two Tesla T4 GPUs for 5 epochs with a learning rate of 2e-4 and a batch size of 16. The model checkpoint with the highest validation score in the 5 epochs was selected and used to evaluate the Test Set.

\section{Error Analysis}\label{apd:error_anaysis}
The error analysis details on a sample set of mispredictions by the best baseline model (PubMedBERT) is given in this section. The analysis was done manually for about 100 mispredictions that were sampled.This could be used for further research to improve the models/methods on the dataset.

\begin{itemize}
    \item \textbf{Multi-hop reasoning}: It was observed that the model often mispredicted the questions related to the cause of an event (diagnosis) and the right course of action (treatment) in a given medical situation. Such questions typically require information on multiple symptoms, ailments, and treatments to select the most appropriate choice. This multiplicity of information is not likely to be present in one passage, possibly the reason for the mispredictions.
    \item \textbf{Incorrect context passages}: It is observed that inadequate contexts from the retriever are also major contributors to the mispredictions.
    \item It is found that the models mispredicted the questions requiring arithmetic reasoning. This is in line with the observations in \citep{Dua2019} on BERT-based models.
\end{itemize}

\section{Result \& Discussion}

In this section, the results from the evaluation of the methods discussed in section \ref{apd:exp} are presented.

\begin{itemize}
    \item It is observed that PubMedBERT performs better than other models in all the categories. This aligns with the results from \citep{Gu2020} where PubMedBERT surpasses all other biomedical models in the majority of BLURB  tasks. Examples of correct and incorrect predictions of the model is presented in Table \ref{tab:predictions}
    
    \item PubMedBERT is followed by SciBERT (mix domain pretraining) and BioBERT (continual pretraining) in accuracy. From this result, it can be inferred that the model's performance decreases with a decrease in domain specificity of the models and external knowledge sources.
    \item It is observed that there is an insignificant improvement in the model's performance when Wikipedia is used as context compared to without context results, and the model variants trained on PubMed, which have a 4-7\% improvement in the performance. This can be attributed to the domain specificity of the external knowledge source required by the dataset. The majority of the reasoning types (Diagnosis, treatment, etc.) mentioned in \ref{apd:re_type} require domain expertise as these questions are intended for post-graduate medical students.
    \item The subject wise accuracies of the top PubMedBERT model is presented in Table \ref{tab:subject_wise}
\end{itemize}

% \begin{table}[!ht]
% \small
% \centering
% \begin{tabular}{lccccc}
% \toprule
%  & {\bf Train} & {\bf Test} & {\bf Dev} & {\bf Total} \\
% \midrule
% Question \#  &  1,82,822 & 6,150  & 4,183  & 1,93,155 \\
% Vocab & 94,231& 11,218 & 10,800 & 97,694 \\
% Max Q tokens & 220 & 135 & 88 & 220  \\
% Max A tokens & 38 & 21 & 25& 38 \\
% Max E tokens & 3,155 & 651& 695& 3,155 \\
% Avg Q tokens & 12.77 & 9.93 & 14.09& 12.71 \\
% Avg A tokens & 2.69& 2.58& 3.19& 2.70 \\
% Avg E tokens & 67.52 & 46.54 & 38.44& 66.22 \\
% \bottomrule
% \end{tabular}
% \caption{MedMCQA dataset statistics}
% \label{tab:data_split}
% \vspace{-2ex}
% \end{table}

\begin{table}[!ht]
\small
\centering
\begin{tabular}{l|ccccc}
\toprule
 {\bf Subject Name} & {\bf Test} & {\bf Dev} \\
\midrule
Anaesthesia  &  0.47 & 0.26  \\
Anatomy      &  0.40 & 0.39  \\
Biochemistry &  0.48 & 0.49  \\
Dental       &  0.43 & 0.36  \\
ENT          &  0.47 & 0.52  \\
FM           &  0.48 & 0.35  \\
O\&G         &  0.54 & 0.39  \\
Medicine     &  0.49 & 0.47  \\
Microbiology &  0.50 & 0.44  \\
Ophthalmology&  0.60 & 0.51  \\
Orthopaedics &  -    & 0.33  \\
Pathology    &  0.53 & 0.46  \\
Pediatrics   &  0.39 & 0.45  \\

Pharmacology &  0.46 & 0.46  \\
Physiology   &  0.47 & 0.47  \\
\rowcolor{rowgray}
\textbf{Psychiatry}   &  \textbf{0.67} & \textbf{0.56}  \\
Radiology    &  0.42 & 0.31  \\
Skin         &  0.50 & 0.29  \\
PSM          &  0.44 & 0.35  \\
Surgery      &  0.50 & 0.43  \\
Unknown      &  0.44 & 1.0 \\

\bottomrule
\end{tabular}
\caption{Fine-grained evaluation per medical subject in test and dev set}
\label{tab:subject_wise}
\vspace{-2ex}
\end{table}

% \begin{table*}[t!]
% \small
%     \centering
%     \resizebox{0.5\textwidth}{!}{%
%     \begin{tabular}{@{}l|r@{ }r@{ }r|r@{}r |r@{}r}
%     \toprule
%     {} & \multicolumn{2}{c|}{\bf Multilingual} & \multicolumn{2}{c|}{\bf Cross-lingual} \multicolumn{2}{c}{\bf Multilingual}\\
%     \midrule
%     \bf Language & \multicolumn{1}{c}{\bf Train} &  \multicolumn{1}{c}{\bf Dev} &  & \multicolumn{1}{c}{\bf Train} & \multicolumn{1}{c}{\bf Dev} & \multicolumn{1}{c}{\bf Train} & \multicolumn{1}{c}{\bf Dev} \\
%     \midrule
%     Albanian & 565 & 185 & 1,194 &  311 & 1,194 &  311\\
%     Arabic           &      755 &   562 &            755 &  755 & 1,194 &  311  \\\midrule
%     Combined       &   2,672 & 13,510 &        755 & 755  & 1,194 &  311 \\
%     \bottomrule
%     \end{tabular}
%     }
%     \caption{Number of examples in the data splits based on the experimental setup. 
%     }
%     \label{tab:datasplits}
% \end{table*}

\begin{table}[t!]
\small
    \centering
    \resizebox{0.45\textwidth}{!}{%
    \begin{tabular}{@{}l|r@{ }r@{ }|r@{ }r@{ }|rrr@{ }}
    \toprule
    {} & \multicolumn{2}{c|}{\bf w/o  
Context} & \multicolumn{2}{c|}{\bf Wiki} & \multicolumn{2}{c}{\bf PubMed}\\
    \midrule
    \bf Model & \multicolumn{1}{c}{\bf Test} &  \multicolumn{1}{r|}{\bf Dev} & \multicolumn{1}{c}{\bf Test} & \multicolumn{1}{r|}{\bf Dev} & \multicolumn{1}{c}{\bf Test} & \multicolumn{1}{c}{\bf Dev}\\
    \midrule
    Bert$_{Base}$        & 0.33  &   0.35 &        0.33 &  0.35 &        0.37 &  0.35\\
    BioBert           &           0.37  &   0.38 &            0.39 &  0.37   &        0.42 &  0.39\\
    SciBert        &       0.39  & 0.39 &        0.38 &  0.39 &        0.43 &  0.41\\
      \rowcolor{rowgray}

    PubMedBERT        &  \textbf{0.41}  & \textbf{0.40}  &        \textbf{0.42}  &  \textbf{0.41} &        \textbf{0.47} &  \textbf{0.43}\\
    \bottomrule
    \end{tabular}
    }
    \caption{Performance of all baseline models in accuracy (\%) on MedMCQA test-dev set
    }
    \label{tab:acc}
\end{table}

\section{Conclusion}

In this work, MedMCQA, a new large-scale, Multi-Choice Question Answering (MCQA) dataset, is presented, which requires a deeper domain and language understanding as it tests the 10+ reasoning abilities of a model across a wide range of medical subjects \& topics. It is demonstrated that the dataset is challenging for the current state-of-the-art methods and domain-specific models, with the best baseline achieving only 47\% accuracy. It is expected that this dataset would facilitate future research in this direction.

\section*{Institutional Review Board (IRB)}
This research does not require IRB approval.
\bibliography{jmlr-sample}
\appendix
\section{Topic Distribution}\label{apd:first}
\begin{figure}[!ht]
\centering
  \includegraphics[width=7.5cm]{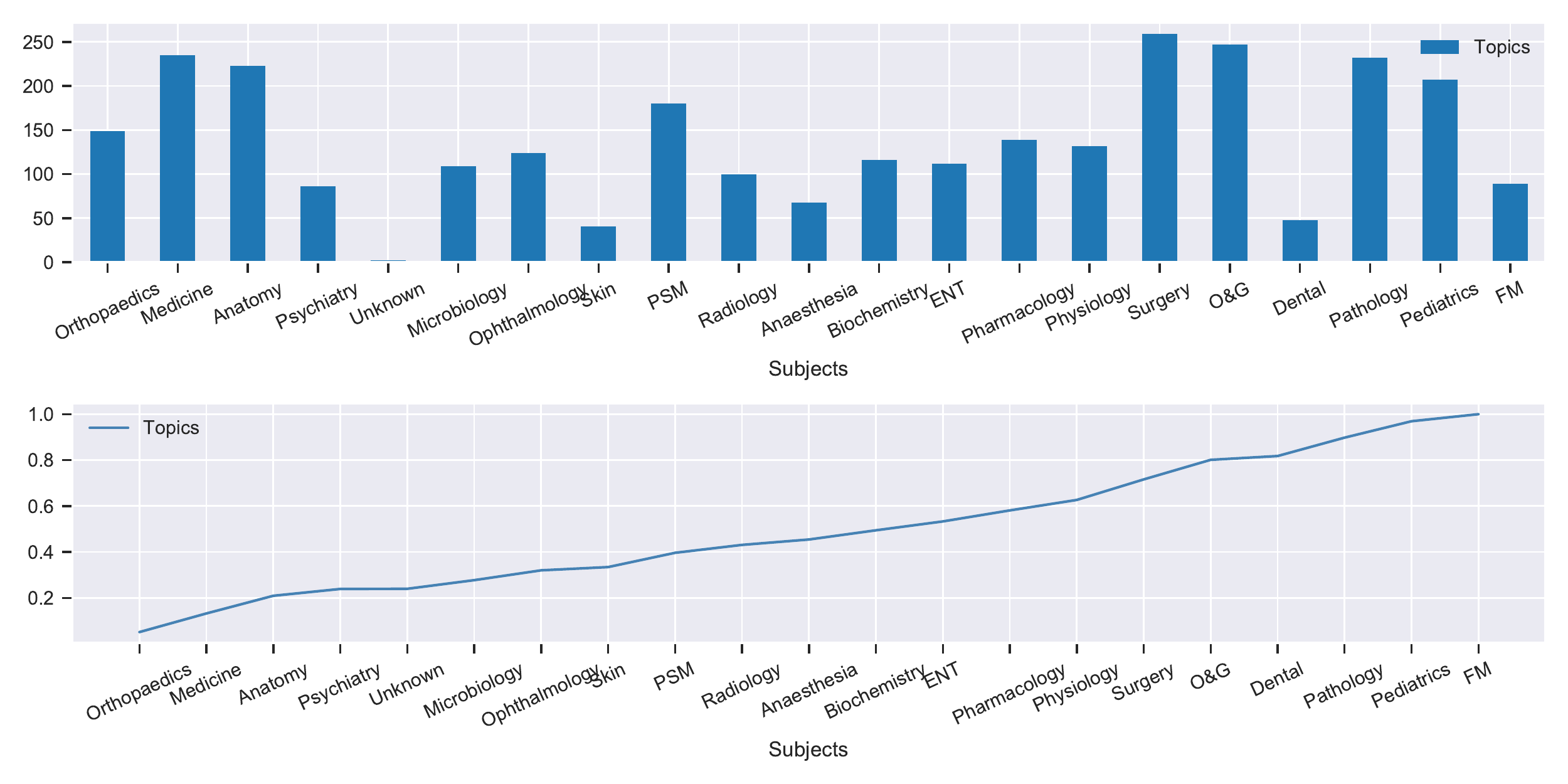}
  \caption{ \footnotesize 
Distribution of topics per subject \& Cumulative Frequency Graph for MedMCQA dataset. }
  \label{fig:topic_per_sub}
\end{figure}

\pagebreak
\section*{Predictions from the best model}
\label{tab:predictions}
\subsection{Correct Predictions}
\begin{table}[htbp]
\small
\centering
\begin{tabular}{|p{18em}|c|c|c|c|b{5em}|c|}
    \hline
        {\bf Question} & {\bf Correct option} & {\bf Options} & {\bf Prediction} \\ \hline
         A 10-year-old boy is having sensory neural deafness. He showed no improvement with conventional hearing aids. Most appropriate management is: & D & \vbox{\hbox{\strut}\hbox{\strut A. Bone conduction hearing aids}\hbox{\strut B. Fenestration}\hbox{C. Stapes fixation}\hbox{D. Cochlear implant}} & D \\ \hline
         Meralgia paraesthetica is due to the involvement of: & A & \vbox{\hbox{\strut}\hbox{\strut A. Lateral cutaneous nerve of the thigh}\hbox{\strut B. Sural nerve}\hbox{\strut C. Medial cutaneous nerve of the thigh}\hbox{\strut D. Femoral nerve}} & A \\
        \hline
         Xanthenuric acid is produced in metabolism of? & A & \vbox{\hbox{\strut}\hbox{\strut A.Tyrosine}\hbox{\strut B.Glycine}\hbox{\strut C.Methionine}\hbox{D.Tryptophan}} & A \\
        \hline
         A 10 years — old child is brought to the emergency room with seizures of the tonic — clonic type. His mother reports
that these seizures have been occurring for the past 50 minutes. The treatment of choice is. & A & \vbox{\hbox{\strut}\hbox{\strut A.Diazepam}\hbox{\strut B.Phenytoin}\hbox{\strut C. Carbamazepine}\hbox{\strut D.Valproate}} & A \\
        \hline
         Which drug is a selective COX 2 inhibitor? & A & \vbox{\hbox{\strut}\hbox{\strut A. Celecoxib}\hbox{\strut B.Acetaminophen}\hbox{\strut C.Ketorolac}\hbox{\strut D.Aspirin}}&A \\
        \hline
\end{tabular}
\end{table}

\bigskip
\subsection{Incorrect Predictions}
\bigskip
\onecolumn
\begin{longtable}{|p{20em}|c|c|c|c|b{8em}|c|}
    \hline
        {\bf Question} & {\bf Correct option} & {\bf Options} & {\bf Prediction} \\ \hline
         Drug of choice for American trypanosomiasis is? & D & \vbox{\hbox{\strut}\hbox{\strut A. Miltefosine}\hbox{\strut B. Amphotericin}\hbox{C. Amphotericin}\hbox{D. Amphotericin}} & A \\ \hline
         Which of the following drugs dosage interval should be maximum in a patient with creatinine clearance less than 10, & C & \vbox{\hbox{\strut}\hbox{\strut A. Amikacin}\hbox{\strut B. Rifampicin}\hbox{\strut C. Vancomycin}\hbox{\strut D. Amphotericin}} & B \\
        \hline
         Filgrastrim is used for: & A & \vbox{\hbox{\strut}\hbox{\strut A.Neutropenia}\hbox{\strut B.Anemia}\hbox{\strut C.Polycythemia}\hbox{D.Neutrophilia}} & C \\
        \hline
         A 30 years old male is having prpductive cough with dysnea. Blood gas analysis shows low pa02. Chest x-ray is showing reticulonodular pattern. The causative agent is? & C & \vbox{\hbox{\strut}\hbox{\strut A.Staph aureus}\hbox{\strut B.Pneumococcus}\hbox{\strut P. jerovecii}\hbox{\strut Pseudomonas}} & B \\
        \hline
         A population study showed a mean glucose of 86 mg/dL in a sample of 100 showing normal curve distribution, what percentage of people have glucose above 86 mg/dL? & B & \vbox{\hbox{\strut}\hbox{\strut A. 34}\hbox{\strut B.50}\hbox{\strut C.Nil}\hbox{\strut D.68}}&A \\
        \hline
\end{longtable}

\end{document}